\pdfoutput=1

\documentclass[11pt]{article}

\usepackage[final]{acl}

\usepackage{times}
\usepackage{latexsym}
\usepackage[T1]{fontenc}
\usepackage[utf8]{inputenc}
\usepackage{microtype}
\usepackage{inconsolata}
\usepackage{graphicx}
\usepackage{booktabs}
\usepackage{subfigure}
\usepackage{listings}
\usepackage{tabularx}
\usepackage{amsmath}
\usepackage{amssymb}
\usepackage{mathtools}
\usepackage{amsthm}
\usepackage{hyperref}
\usepackage[normalem]{ulem}
\usepackage{float} 

\lstdefinelanguage{Prompt}{
    basicstyle=\ttfamily,
    backgroundcolor=\color{gray!10},
    frame=none,
    breaklines=true,
    moredelim=[s][\textbf]{\{}{\}}
}

\begin{document}
\title{Model Consistency as a Cheap yet Predictive Proxy for LLM Elo Scores}

\author{Ashwin Ramaswamy\\
  Independent \\
  \texttt{ashwinsrama@gmail.com} \\\And
  Nestor Demeure \\
  Lawrence Berkeley\\National Laboratory \\
  \texttt{ndemeure@lbl.gov} \\\And
  Ermal Rrapaj \\
  Lawrence Berkeley\\National Laboratory \\
  \texttt{ermalrrapaj@lbl.gov} \\}

\maketitle
\begin{abstract}
New large language models (LLMs) are being released every day. Some perform significantly better or worse than expected given their parameter count.  
Therefore, there is a need for a method to independently evaluate models.  
The current best way to evaluate a model is to measure its Elo score by comparing it to other models in a series of contests—an expensive operation since humans are ideally required to compare LLM outputs.  
We observe that when an LLM is asked to judge such contests, the consistency with which it selects a model as the best in a matchup produces a metric that is 91\% correlated with its own human-produced Elo score.  
This provides a simple proxy for Elo scores that can be computed cheaply, without any human data or prior knowledge. 
\end{abstract}

\section{Introduction}
\label{Introduction}

The rapid proliferation of large language models (LLMs) makes it increasingly difficult to assess which models are best suited for specific tasks, especially given that some models outperform or underperform relative to their parameter count \cite{hong2023diminishingreturnsmaskedlanguage, grattafiori2024llama3herdmodels}. This challenge has motivated the need for independent, scalable model evaluation methods.

Measuring the intelligence of LLMs has been pursued through various approaches \cite{peng2024surveyusefulllmevaluation}, from task-specific benchmarks \cite{HernndezOrallo2017EvaluationIA} to natural language generation analysis \cite{10.3115/1073083.1073135}. The most prominent approach is human-based evaluation, exemplified by the LMSYS Chatbot Arena Leaderboard \cite{chiang2024chatbot}, where human annotators compare responses from different models to generate Elo scores. However, this process is subjective and, due to its inherent dependence on humans, cannot scale with the increasing number of newly released models that need to be evaluated to keep the Elo scores updated.

To address scalability, researchers have turned to LLMs-as-Judges, replacing human evaluators with high-performing LLMs. While this approach eliminates the need for human data, it introduces biases including position bias \cite{shi2024judgingjudgessystematicstudy}, verbosity bias \cite{saito2023verbositybiaspreferencelabeling}, and self-enhancement bias \cite{xu2024prideprejudicellmamplifies}. More critically, LLM judges show poor alignment with human preferences \cite{liu2025aligninghumanjudgementrole}, limiting their viability as replacements for human evaluators.

Objective tests like MMLU \cite{hendrycks2021measuringmassivemultitasklanguage} and MedQA \cite{jin2020diseasedoespatienthave} offer another approach, evaluating domain-specific knowledge without human input. While these benchmarks correlate with human preferences \cite{holliday2024conditionalmodalreasoninglarge}, they fail to assess style or complex task performance, and risk overestimating models trained on benchmark data \cite{zhou2023dontmakellmevaluation}.

Despite numerous alternatives, the Elo score method remains the gold standard \cite{quan2025codeelobenchmarkingcompetitionlevelcode, gonzálezbustamante2024textclassbenchmarkcontinuouselo, wu2023stylesubstanceevaluationbiases} because it captures both stylistic preferences and diverse task capabilities that are difficult to quantify. However, its inherent flaws \cite{singh2025leaderboardillusion} warrant consideration of other properties of LLMs that can approximate the Elo score.

In the following sections, we will introduce our proposed method and its computation, as well as a set of experiments demonstrating the high correlation of our proposed metric with Elo scores.
We also provide observations on the type of data that is best suited for producing such results, concluding with perspectives on how our preliminary findings could be further improved.
\section{Proposed Method}  
\label{Proposed Method}  


We propose that an LLM's Consistency—quantified by the variance measured in its pairwise decisions when acting as a judge—can serve as a proxy for measuring its own Elo score. This does so without relying on human evaluation, prior knowledge, or susceptibility to traditional data leakage issues and independent judge LLM biases.

\subsection{Consistency Formula}


Given two models, \(m_i\) and \(m_j\), and a model serving as a judge \(m_\text{judge}\), running a \textbf{contest} between that pair of models consists of asking the same question to both models and having the judge pick the best answer (see Section~\ref{sec:elicitation_process} for details on the process of running an individual contest). A \textbf{matchup} \(ij\) is the set of all contests between models \(m_i\) and \(m_j\), evaluated by the same judge \(m_\text{judge}\). 

If the judge's aggregated preferences for a matchup yields a probability of winning \(p_{ij}\) to \(m_i\), we can model the matchup with a Bernoulli distribution of parameter \(p_{ij}\) and variance \(p_{ij} \cdot (1 - p_{ij})\). This variance is inversely correlated with the consistency with which the judge selects a model as the winner: it approaches \(0.25\) if both models are equally likely to win and tends toward \(0\) if the same model's answer is always preferred.  


Given a set of matchups \(M\) between models, if we denote \(n_{ij}\) as the number of contests between models \(m_i\) and \(m_j\), we can compute the average variance across matchups using Equation~\ref{eq:wvar}.  

\begin{equation}  
\overline{\text{Var}}(m_\text{judge}, M) = \frac{ \sum_{ij \in M}{n_{ij} \cdot p_{ij} \cdot (1 - p_{ij})} }{ \sum_{ij \in M}{n_{ij}} }  
\label{eq:wvar}  
\end{equation}  


This average variance can be used to measure a judge's consistency across the set of matchups \(M\).
We can further rescale the variance to make it easier to interpret, producing the \textbf{Consistency score} formula in Equation~\ref{eq:consistency}, yielding a number between \(0\) (the judge is entirely random) and \(1\) (the judge always picks the same models as winners in their respective matchups). 

\begin{equation}  
\text{Consistency}(m_\text{judge}, M) = 1 - 4 \cdot \overline{\text{Var}}(m_\text{judge}, M)
\label{eq:consistency}  
\end{equation}  


Hypothesizing that an LLM with a high Elo score, as typically determined over a number of contests, would itself be good at judging such contests, and that a good judge is expected to have low variance, being consistent in its answers, we explored using the Consistency score as a proxy for a model's Elo score.  

Consistency scores have the benefit of not requiring any human feedback or prior knowledge of the task (we do not need to evaluate whether the judged models were \emph{correct} in their answers),
being cheap and easy to compute (given a number of preexisting answers to questions, we only need to run the judge over each contest, selecting winners),
and being applicable to a wide array of tasks, including less well-defined ones where no clear-cut answer may be available (such as complex textual tasks like translation).

To further motivate for our experimentation, we saw that \citet{wen2025unsupervisedelicitationlanguagemodels} employs a logical consistency function across many data points to maximize mutual predictability of labeling, implying that the consistency with which a label is estimated gives insight into the capabilities of an LLM for downstream tasks.
\subsection{Preference Elicitation}
\label{sec:elicitation_process}

We employ a two-stage chain-of-thought prompting process to elicit preferences from LLMs judging a given contest.

First, the judge LLM is presented with the original user question and both model responses. The judge is then prompted to generate a paragraph of reasoning explaining its preference based on a predefined set of evaluation criteria (see the prompt in Appendix~\ref{prompt:reasoning}).

Second, the judge LLM receives the generated reasoning and is prompted to extract a preference from it, selecting either response `1', `2', or `0' to indicate a tie (see the prompt in Appendix~\ref{prompt:picking}).

This two-stage approach has proven necessary because single-pass prompting often led judge LLMs with lower Elo scores to misunderstand the task—either attempting to answer the question themselves or elaborating on the prompt rather than providing a preference.
By first generating a justification, we ensure that the judge LLM adheres to the evaluation criteria, reducing preference selection errors.

Additionally, allowing ties prevents artificially inflated variance caused by forcing judges to make arbitrary selections in cases where neither response is clearly preferable. This approach aligns with the LMSYS dataset format, ensuring compatibility and consistency in evaluation.

For each model \(m_i\) competing against a model \(m_j\), the probability of \(m_i\) winning against \(m_j\) (denoted \(p_{ij}\)) is computed using Equation~\ref{eq:win_proba},
based on the number of contests between the two models (\(n_{ij}\)), the number of wins of model \(m_i\) (\(\text{Wins}_{ij}\)), and the number of ties (\(\text{Ties}_{ij}\), with each tie counted as half a win).

\begin{equation}
    p_{ij} = \frac{ \text{Wins}_{ij} + 0.5 \cdot \text{Ties}_{ij} }{ n_{ij} }
    \label{eq:win_proba}
\end{equation}

Consistent with findings from \citet{shi2024judgingjudgessystematicstudy}, we observe a strong positional bias toward the first response in preference elicitation (less pronounced with higher Elo models).
To mitigate this bias, we submit an equal number of contests for each possible ordering of models (both \(ij\) and \(ji\)), thus avoiding artificially increasing the Consistency score of a judge by always placing a judged model in the first position within its contests.


\section{Experiments}
\label{Experiments}

\subsection{Experimental Setup}
\label{sec:result_setup}

We employed 24 LLMs, selected to cover a diverse range of Elo scores (listed in Table~\ref{table:judges} in section ~\ref{sec:matchup_tables} of the Appendix). These scores range from 1008 to 1315, based on the most recent available LMSYS Chatbot Arena Leaderboard dataset. Each judge LLM was used to evaluate and select preferences between responses generated by other LLMs.

The LMSYS Elo scores are somewhat noisy; Elo scores and model rankings evolve as new models and contests are introduced in the LMSYS Chatbot Arena. As such, we do not expect to perfectly fit any frozen set of Elo scores.


To test our hypothesis, we used contests from the \textit{arena-human-preference-55k} dataset \cite{chiang2024chatbot}, which contains a collection of unique prompts crowdsourced by real human users, paired responses to each prompt from two competing LLMs, along with the identities of the competing LLMs. Although this dataset also includes human preference annotations, these were not used in our study. We used this dataset because it provided us exactly what we needed: pairs of responses from LLMs and a wide range of LLMs responding to prompts that would have minimal overlap with the judge LLMs being evaluated. 

From this dataset, we extracted a subset of 2800 question-response pairs involving 35 different LLMs across 140 different matchups, as detailed in Section~\ref{sec:matchup_tables} of the Appendix.
Models were selected so that each pair of models participated in at least 20 contests and that most of these models differed from our 24 judge LLMs\footnote{The code and data used in the experiment is available at \href{https://github.com/ashwinsrama/LLM_Consistency}{https://github.com/ashwinsrama/LLM\_Consistency}}.

\subsection{Correlation with Elo Score}
\label{sec:result_correlations}

\begin{figure}[h!]
    \centering
    \includegraphics[width=1\linewidth]{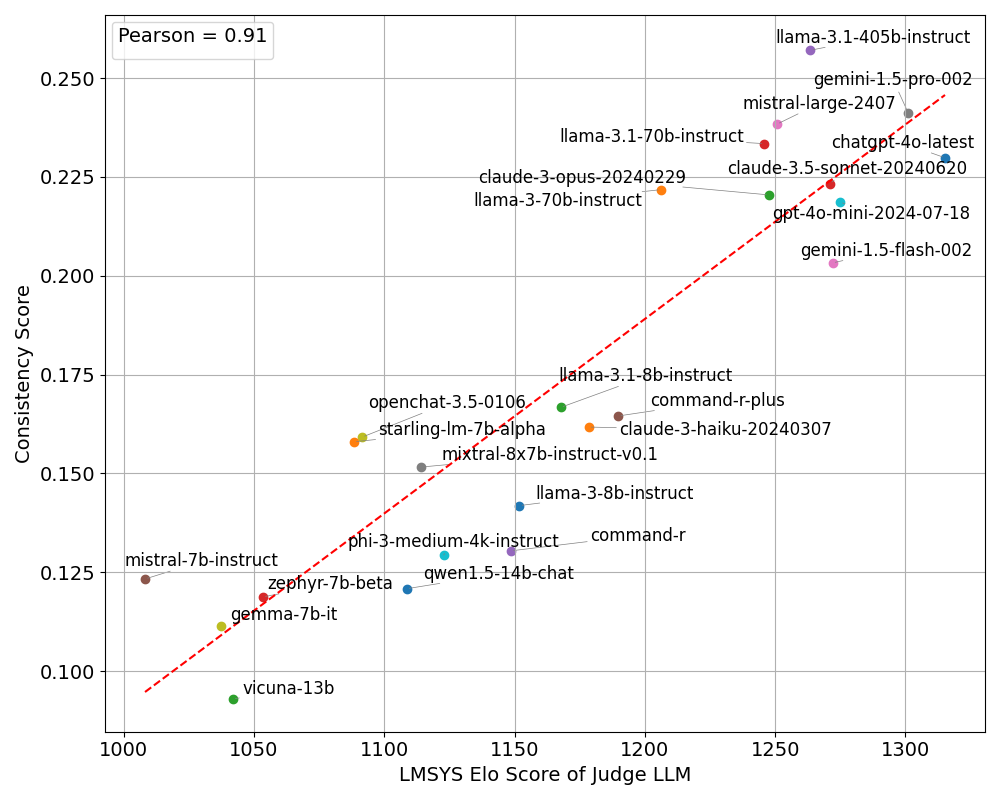}
    \caption{Correlation plot between the Elo score and Consistency score of each judge.}
    \label{fig:correlation}
\end{figure}

Plotting the Consistency score (defined in Equation~\ref{eq:consistency}) of each judge LLM over the full set of matchups against the Elo scores of the judge LLMs (based on the LMSYS Chatbot Arena Leaderboard) yielded a Pearson correlation coefficient of \(0.91\), as depicted in Figure~\ref{fig:correlation}.

We found similarly high correlations when examining different sets of judge models and using different contests to measure Consistency scores, provided that the models in contest matchups had sufficiently large Elo differences (see Section~\ref{sec:impact_of_data}) and the judges spanned a diverse range of Elo scores.

While the distribution of judge LLMs was chosen to be uniform across Elo scores, there appear to be roughly three clusters when plotted against Consistency scores: low-performing models (lower part of the plot), medium-performing models, and high-performing models (upper part of the plot).

We found that the correlation between Elo score and Consistency score is significantly lower within each cluster: \(0.74\), \(0.76\), and \(0.12\) for the low-, medium-, and high-performing clusters, respectively (see Appendix~\ref{sec:cluster_correlations} for details).
While this is in part due to the small size of these clusters, it suggests that our metric is capable of distinguishing broad differences in model capabilities but is not yet refined enough to differentiate, for example, between the five best models available.

Ranking models by Consistency scores places them on average within \(2.8\) positions of their Elo ranking, while predicting a model's Elo score based on its Consistency score, using simple linear regression, results in a mean absolute error of \(35.2\) Elo points.

\subsection{Impact of the Contest Data}
\label{sec:impact_of_data}

\begin{figure}[h]
    \centering
    \includegraphics[width=1\linewidth]{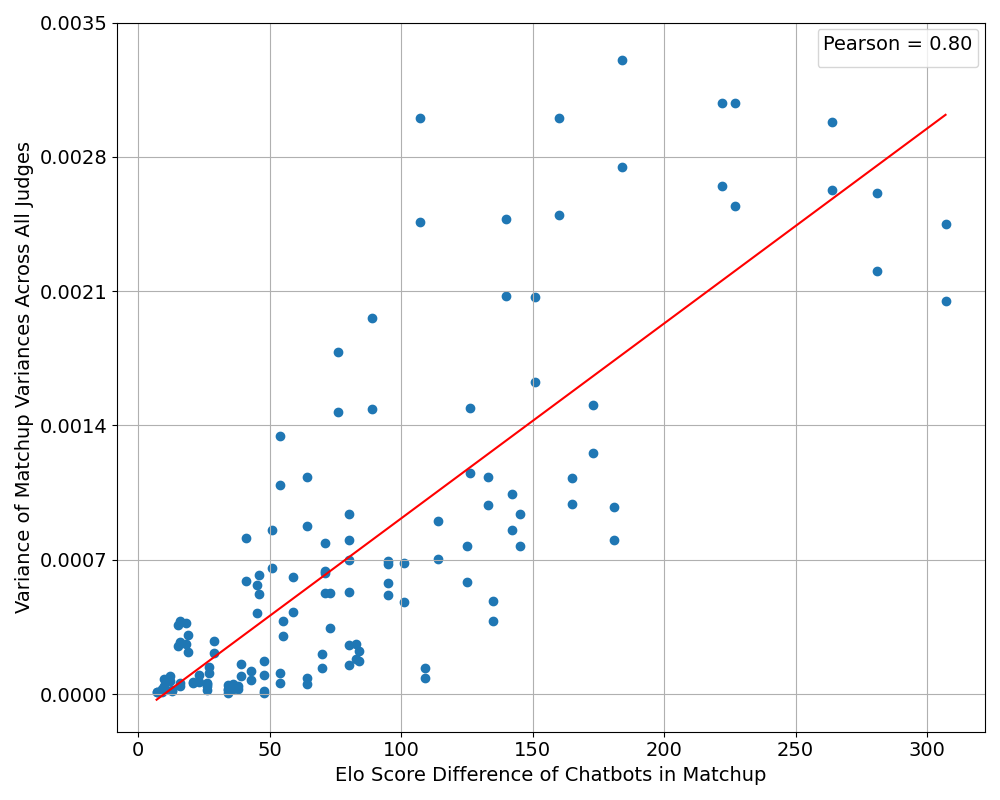}
    \caption{Correlation plot between the difference in Elo scores in a given matchup of two models and the variance of matchup variance computed across all judge LLMs for that matchup.}
    \label{fig:variance_vs_elodiff}
\end{figure}

We further observe a strong and positive correlation when computing the variance (per Equation~\ref{eq:wvar}) of judges on \emph{individual} matchups and then plotting the variance of those variances against the difference in Elo scores between the two models in the matched pair (see Figure~\ref{fig:variance_vs_elodiff}).

This is somewhat intuitive: if the two models in a matchup have similar Elo scores, then no judge manages to reliably differentiate them, leading to low consistency across all judges and correspondingly low variance between judges.
However, pairs with a larger Elo difference are easier to discriminate, allowing strong judges to excel while weak judges remain unable to consistently pick the better response, thereby increasing the variance on that matchup.
This suggests that pairs of models with large Elo differences are the most effective for identifying consistent judges.

\begin{figure}[h]
    \centering
    \includegraphics[width=1\linewidth]{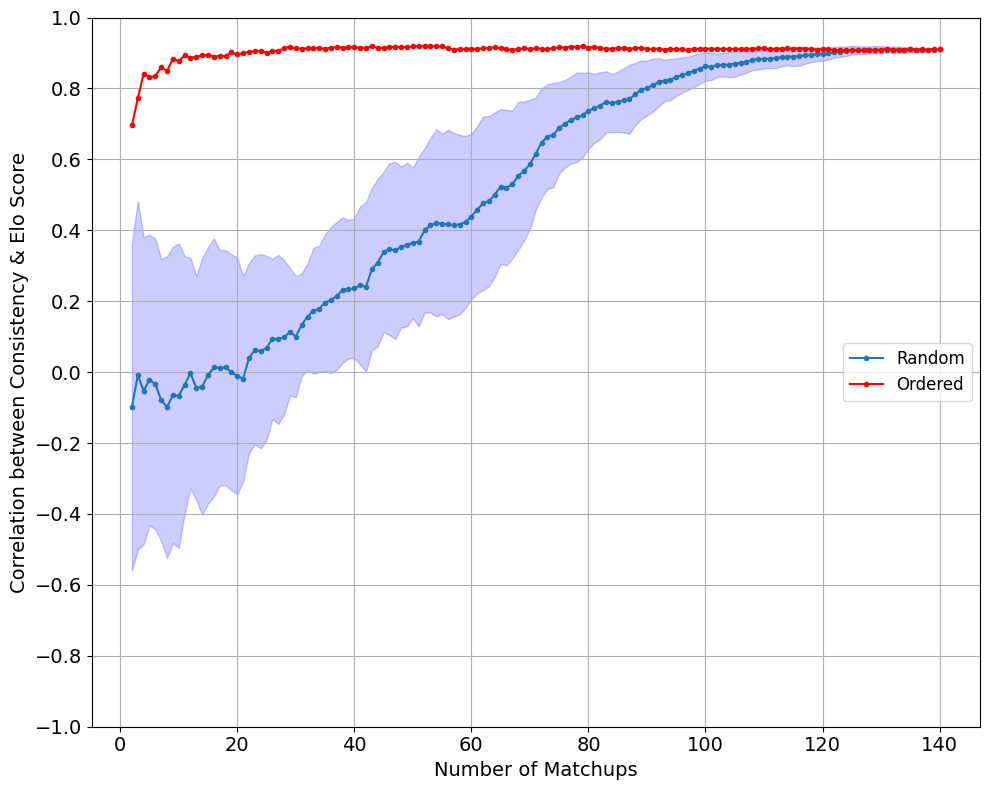}
    \caption{Convergence of the Pearson correlation coefficient between Consistency and Elo scores as the number of matchups used to compute the Consistency scores increases.
    The blue line is computed over 25 random matchup orders, while the red line is produced by adding the matchups with the highest Elo differences first.}
    \label{figure:convergence}
\end{figure}

Building on this insight, we examined the convergence of our metric by measuring the correlation between Consistency and Elo scores as we increased the number of matchups (see Figure~\ref{figure:convergence}), either randomly (blue line, averaged over 25 random matchup orders) or by adding matchups with larger Elo differences first (red line).

Both lines converge to a \(0.91\) correlation, as they eventually use the same matchups. However, we found that most of the convergence can be achieved using just the 10 matchups with the highest Elo differences, reaching a correlation of \(0.88\) (see Figure~\ref{fig:correlation_top10matchups} in the Appendix). The judges yield a significantly higher Consistency score compared to the previous correlation plot, because the matchups are easier for judges to be consistent upon. Additionally, we found that using just 30 matchups is sufficient to extract the full information, reaching a \(0.91\) correlation.


This highlights the importance of the contest data in computing the metric.
Matchups between models with widely different Elo scores are necessary to extract information about the judge's ability to discriminate between models, an observation consistent with \citet{gao2024reevaluatingautomaticllmranking}.


Regarding the number of matchups to be used, it is interesting to note that the sorted matchups convergence curve is essentially monotonous, suggesting that, while adding matchups with lower Elo differences does dilute consistency (as seen in Figure~\ref{fig:correlation_top10matchups}), it appears to be an additive operation: adding information without harming the correlation of the metric with Elo scores.
This means that additional matchups can be incorporated as long as matchups with high Elo differences are prioritized.

\section{Conclusion}

We introduce Consistency scores as a simple, scalable metric for estimating LLM intelligence by measuring a model's consistency in preferring the same winner in repeated pairwise matchups. This score strongly correlates with LMSYS Elo scores ($r=0.91$), achieving a mean prediction error of just \(35.2\) Elo points—all without requiring human input or prior knowledge of the models.

Our method is efficient to apply: evaluating a new model's Elo score only requires running pairwise comparisons on existing answer sets. It generalizes well across tasks, including subjective ones, though its effectiveness improves when contest data features models with widely differing abilities. 

Overall, we believe that Consistency scores present a promising avenue for measuring model intelligence due to their ease of measurement and the strong results observed in this initial exploration. This approach offers a promising direction for automatic LLM evaluation, with future work focused on refining rankings among top-tier models. We hope that further refinements in this direction will lead to a cost-effective and reliable metric for evaluating model intelligence across a diverse range of traditionally challenging tasks.
\section*{Limitations}

While our work is novel and demonstrates a correlation between judgment consistency and intelligence, there are shortcomings that practitioners should consider when using this metric. 

Primarily, the metric lacks the ability to distinguish between top-performing models. As depicted in Appendix~\ref{sec:cluster_correlations}, the highest cluster yields a Pearson correlation of 0.12. However, as research labs and companies investigate smaller models---for tasks such as low-power LLM inference, knowledge distillation, and testing fine-tuning algorithms---even the strong correlation within the lower clusters of LLMs makes the metric useful.

Additionally, while the Consistency score is not able to distinguish between models that are very close in Elo score, one should note that Elo scores are a noisy metric, because it is inherently subjective and can change from day-to-day based on the number of human preferences elicited. In fact, the Elo score for each model represented in the LMSYS Chatbot Arena Leaderboard comes with a 95\% confidence interval that can drastically change the range of permissible Elo scores and relative ranking of each model.

Looking beyond these limitations, we identify several areas for potential improvement on our existing work that warrant attention.

First, curating dedicated contest sets optimized for information extraction could enhance performance beyond the essentially random data utilized in this study. 

Second, a weighting formula could be designed to account for Elo differences in matchups, as pairs with smaller differences yield noisier results and might merit downweighting.

Additionally, prompt engineering may unlock better performance from stronger models, which might currently be constrained by imprecise instructions.

Finally, a more sophisticated uncertainty model leveraging broader contest data could improve robustness. Such a model should incorporate patterns where consistently high-performing models across diverse contests demonstrate greater reliability than those with variable performance.
\section*{Acknowledgements}

We thank the CBorg group at the Lawrence Berkeley National Laboratory (LBNL) for providing us access to proprietary models and the CS Division at LBNL for funding our use of these models.

We would also like to thank the National Energy Research Scientific Computing Center (NERSC) for access to the Perlmutter Supercomputer, which was used to run inference on some of the models.

This research used resources from the National Energy Research Scientific Computing Center (NERSC), a U.S. Department of Energy Office of Science User Facility located at Lawrence Berkeley National Laboratory, operated under Contract No. DE-AC02-05CH11231.

\bibliography{bibliography}
\clearpage
\appendix
\section{Appendix}

%

\subsection{Prompts Used}
\label{sec:prompts}

\subsubsection{Reasoning Prompt}
\label{prompt:reasoning}

{
\footnotesize 
\begin{lstlisting}[language=Prompt]
Please act as an impartial judge and evaluate the quality of the responses provided by two AI assistants to the user question displayed below.  

[User Question]
{PROMPT}

[The start of Answer #1]
{ANSWER1}
[The end of Answer #1]

[The start of Answer #2]
{ANSWER2}
[The end of Answer #2]

Your task is to determine which of the two answers is better, based on the following criteria:
- Choose the response that follows the user's instructions and answers the user's question better.
- Evaluate based on helpfulness, relevance, accuracy, depth, creativity, and level of detail of the responses.
- Do not allow the length or order of the responses to influence your evaluation.

In 1 paragraph, write out your thought process and reasoning for which answer better meets the criteria. Do not actually answer the original question yourself. You are only choosing between the two provided answers based on how well they meet the specified criteria. If both answers are comparable such that you have no preference, then explain that.
\end{lstlisting}
}

\subsubsection{Preference Prompt}
\label{prompt:picking}

{
\footnotesize 
\begin{lstlisting}[language=Prompt]
Read the following passage: 

[Start of Passage]
{REASONING}
[End of Passage]

Indicate which answer the author thinks is better. Respond with a single integer:
- "1" if the author thinks Answer #1 is better.
- "2" if the author thinks Answer #2 is better.
- "0" if the author has no preference, or thinks the answers are equal.

Your response must be exactly one integer (1, 2, or 0). Do not include any other text/explanation. The author prefers Answer #:
\end{lstlisting}
}

\subsection{Dataset Used}
\label{sec:matchup_tables}

\begin{table}[h!]
\centering
\scriptsize 
\begin{tabularx}{\columnwidth}{Xcr}
\toprule
\textbf{LLM Name} & \textbf{Elo Score} & \textbf{Consistency Score}\\ \midrule
chatgpt-4o-latest & 1315 & 0.230 \\
gemini-1.5-pro-002 & 1301 & 0.241 \\
gpt-4o-mini-2024-07-18 & 1275 & 0.219 \\
gemini-1.5-flash-002 & 1272 & 0.203 \\
claude-3.5-sonnet-20240620 & 1271 & 0.223 \\
llama-3.1-405b-instruct & 1263 & 0.257 \\
mistral-large-2407 & 1251 & 0.238 \\
claude-3-opus-20240229 & 1248 & 0.220 \\
llama-3.1-70b-instruct & 1246 & 0.233 \\
llama-3-70b-instruct & 1206 & 0.222 \\
command-r-plus & 1190 & 0.165 \\
claude-3-haiku-20240307 & 1178 & 0.162 \\
llama-3.1-8b-instruct & 1168 & 0.167 \\
llama-3-8b-instruct & 1152 & 0.142 \\
command-r & 1149 & 0.130 \\
phi-3-medium-4k-instruct & 1123 & 0.129 \\
mixtral-8x7b-instruct-v0.1 & 1114 & 0.152 \\
qwen1.5-14b-chat & 1109 & 0.121 \\
openchat-3.5-0106 & 1091 & 0.159 \\
starling-lm-7b-alpha & 1088 & 0.158 \\
zephyr-7b-beta & 1053 & 0.119 \\
vicuna-13b & 1042 & 0.093 \\
gemma-7b-it & 1037 & 0.111 \\
mistral-7b-instruct & 1008 & 0.123 \\
\bottomrule
\end{tabularx}
\caption{List of the LLMs evaluated in our experiments, sorted by their LMSYS Elo scores in descending order. Consistency scores are computed as per Section~\ref{sec:result_correlations}.}
\label{table:judges}
\end{table}

\begin{table}[h!]
\centering
\scriptsize 
\begin{tabularx}{\columnwidth}{Xr}
\toprule
\textbf{LLM Name} & \textbf{Elo Score} \\ \midrule
gpt-4-1106-preview & 1251 \\
gpt-4-0314 & 1186 \\
gpt-4-0613 & 1162 \\
claude-1 & 1149 \\
mistral-medium & 1148 \\
claude-2.0 & 1132 \\
claude-2.1 & 1118 \\
gpt-3.5-turbo-0613 & 1117 \\
mixtral-8x7b-instruct-v0.1 & 1114 \\
claude-instant-1 & 1111 \\
gpt-3.5-turbo-0314 & 1106 \\
wizardlm-70b & 1106 \\
tulu-2-dpo-70b & 1099 \\
llama-2-70b-chat & 1093 \\
vicuna-33b & 1091 \\
pplx-70b-online & 1078 \\
gpt-3.5-turbo-1106 & 1068 \\
llama-2-13b-chat & 1063 \\
wizardlm-13b & 1059 \\
zephyr-7b-beta & 1053 \\
codellama-34b-instruct & 1042 \\
vicuna-13b & 1042 \\
llama-2-7b-chat & 1037 \\
mistral-7b-instruct & 1008 \\
vicuna-7b & 1005 \\
palm-2 & 1003 \\
koala-13b & 964 \\
mpt-7b-chat & 927 \\
RWKV-4-Raven-14B & 922 \\
oasst-pythia-12b & 894 \\
chatglm-6b & 879 \\
fastchat-t5-3b & 868 \\
stablelm-tuned-alpha-7b & 840 \\
dolly-v2-12b & 822 \\
llama-13b & 799 \\
\bottomrule
\end{tabularx}
\caption{List of LLMs matched in our experiments, sorted by their LMSYS Elo scores in descending order.}
\label{table:matchee}
\end{table}

\begin{table}[!htbp]
\centering
\fontsize{6.5pt}{7.8pt}\selectfont 
\begin{tabularx}{\columnwidth}{XXr}
\toprule
\textbf{First LLM} & \textbf{Second LLM} & \textbf{Difference in Elo Scores} \\ \midrule
gpt-4-0314 & chatglm-6b & 307 \\
fastchat-t5-3b & claude-1 & 281 \\
RWKV-4-Raven-14B & gpt-4-0314 & 264 \\
chatglm-6b & gpt-3.5-turbo-0314 & 227 \\
mpt-7b-chat & claude-1 & 222 \\
RWKV-4-Raven-14B & gpt-3.5-turbo-0314 & 184 \\
gpt-4-0314 & vicuna-7b & 181 \\
fastchat-t5-3b & vicuna-13b & 173 \\
koala-13b & llama-13b & 165 \\
vicuna-33b & gpt-4-1106-preview & 160 \\
tulu-2-dpo-70b & gpt-4-1106-preview & 151 \\
palm-2 & claude-1 & 145 \\
koala-13b & dolly-v2-12b & 142 \\
gpt-4-1106-preview & claude-instant-1 & 140 \\
fastchat-t5-3b & palm-2 & 135 \\
claude-2.1 & gpt-4-1106-preview & 133 \\
chatglm-6b & vicuna-7b & 126 \\
koala-13b & stablelm-tuned-alpha-7b & 125 \\
mpt-7b-chat & vicuna-13b & 114 \\
gpt-3.5-turbo-0613 & mistral-7b-instruct & 109 \\
claude-1 & vicuna-13b & 107 \\
vicuna-7b & gpt-3.5-turbo-0314 & 101 \\
gpt-4-0613 & gpt-3.5-turbo-1106 & 95 \\
llama-13b & oasst-pythia-12b & 95 \\
codellama-34b-instruct & claude-2.0 & 89 \\
gpt-4-0613 & pplx-70b-online & 84 \\
RWKV-4-Raven-14B & vicuna-7b & 83 \\
gpt-3.5-turbo-0314 & gpt-4-0314 & 80 \\
mistral-medium & gpt-3.5-turbo-1106 & 80 \\
gpt-3.5-turbo-0613 & llama-2-7b-chat & 80 \\
palm-2 & mpt-7b-chat & 76 \\
wizardlm-13b & claude-2.0 & 73 \\
oasst-pythia-12b & dolly-v2-12b & 71 \\
koala-13b & oasst-pythia-12b & 71 \\
pplx-70b-online & mistral-medium & 70 \\
gpt-3.5-turbo-0613 & zephyr-7b-beta & 64 \\
wizardlm-70b & codellama-34b-instruct & 64 \\
fastchat-t5-3b & mpt-7b-chat & 59 \\
mistral-7b-instruct & llama-2-13b-chat & 55 \\
stablelm-tuned-alpha-7b & oasst-pythia-12b & 54 \\
gpt-3.5-turbo-0613 & llama-2-13b-chat & 54 \\
llama-2-70b-chat & codellama-34b-instruct & 51 \\
gpt-4-0613 & mixtral-8x7b-instruct-v0.1 & 48 \\
wizardlm-13b & wizardlm-70b & 48 \\
gpt-3.5-turbo-1106 & mixtral-8x7b-instruct-v0.1 & 46 \\
zephyr-7b-beta & mistral-7b-instruct & 45 \\
RWKV-4-Raven-14B & chatglm-6b & 43 \\
llama-13b & stablelm-tuned-alpha-7b & 41 \\
llama-2-70b-chat & claude-2.0 & 39 \\
vicuna-13b & palm-2 & 38 \\
mixtral-8x7b-instruct-v0.1 & pplx-70b-online & 36 \\
mistral-medium & mixtral-8x7b-instruct-v0.1 & 34 \\
llama-2-70b-chat & wizardlm-13b & 34 \\
mistral-7b-instruct & llama-2-7b-chat & 29 \\
vicuna-33b & claude-2.1 & 27 \\
llama-2-13b-chat & llama-2-7b-chat & 26 \\
claude-2.0 & wizardlm-70b & 26 \\
dolly-v2-12b & llama-13b & 23 \\
vicuna-33b & claude-instant-1 & 21 \\
tulu-2-dpo-70b & claude-2.1 & 19 \\
dolly-v2-12b & stablelm-tuned-alpha-7b & 18 \\
llama-2-7b-chat & zephyr-7b-beta & 16 \\
codellama-34b-instruct & wizardlm-13b & 16 \\
gpt-4-0613 & mistral-medium & 15 \\
llama-2-70b-chat & wizardlm-70b & 13 \\
claude-instant-1 & tulu-2-dpo-70b & 12 \\
pplx-70b-online & gpt-3.5-turbo-1106 & 10 \\
zephyr-7b-beta & llama-2-13b-chat & 10 \\
vicuna-33b & tulu-2-dpo-70b & 9 \\
claude-2.1 & claude-instant-1 & 7 \\
\bottomrule
\end{tabularx}
\caption{List of matchup pairs used in our experiments, sorted by their LMSYS Elo score difference.}
\label{table:matchups}
\end{table}

\subsection{Correlation Plots}
\label{sec:cluster_correlations}

\begin{figure}[h!]
    \centering
    \subfigure[Highest cluster]{
        \includegraphics[width=\linewidth]{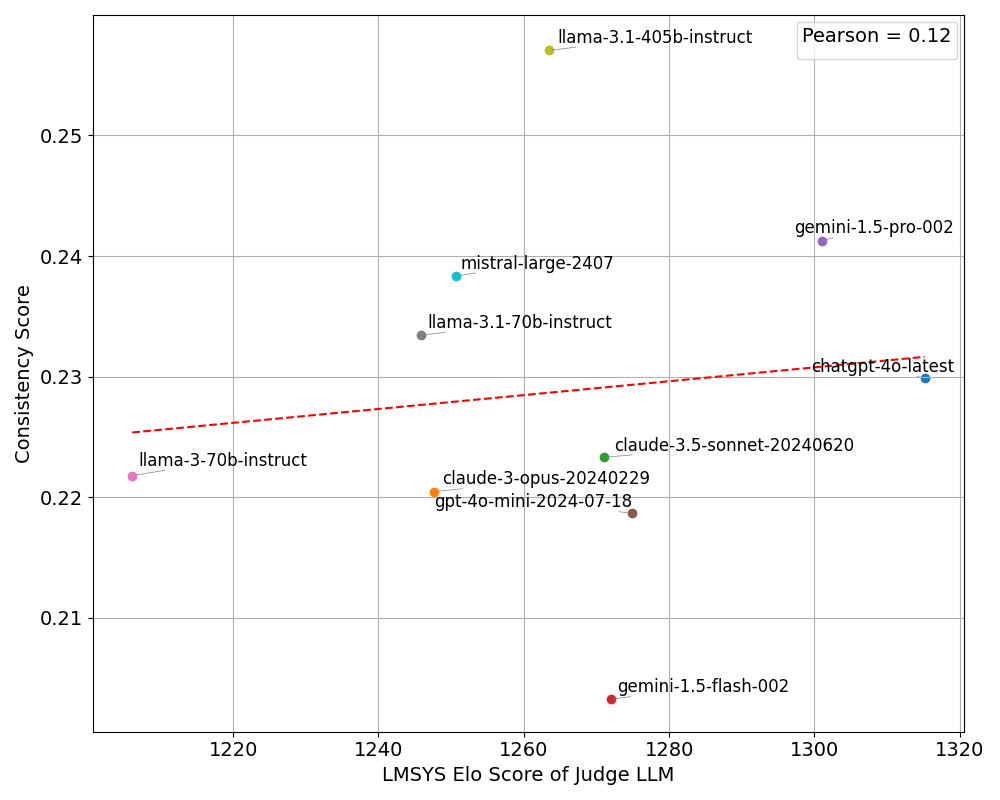}
        \label{fig:correlation_highcluster}
    }
    \subfigure[Middle cluster]{
        \includegraphics[width=\linewidth]{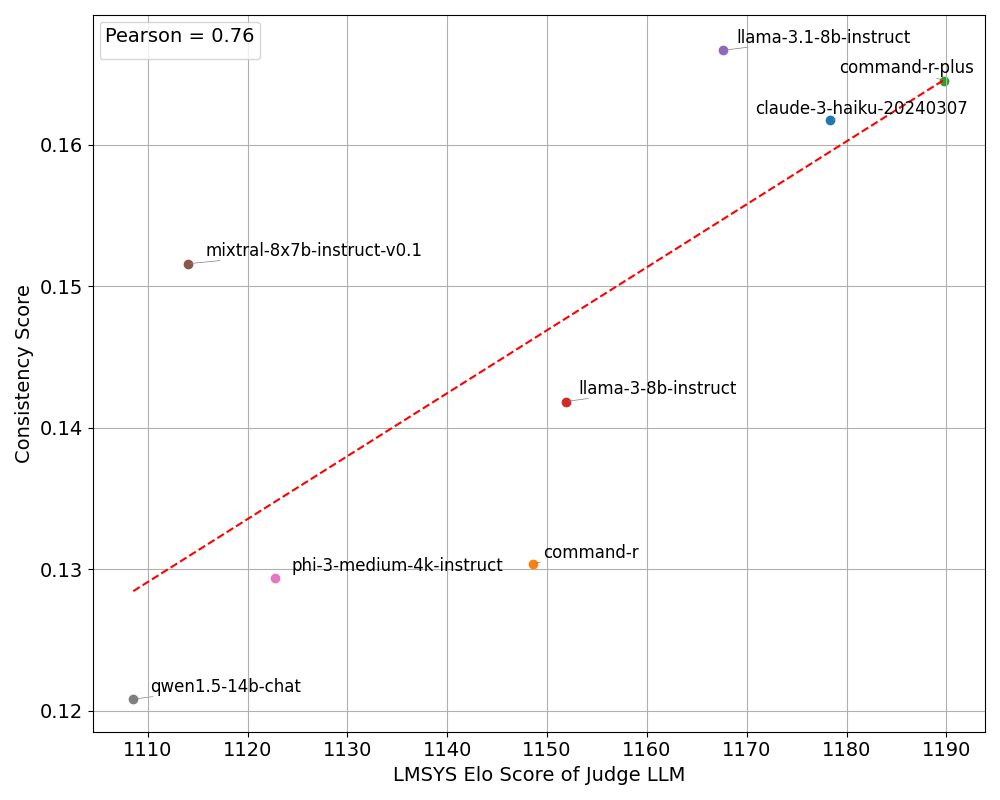}
        \label{fig:correlation_mediumcluster}
    }
    \subfigure[Lowest cluster]{
        \includegraphics[width=\linewidth]{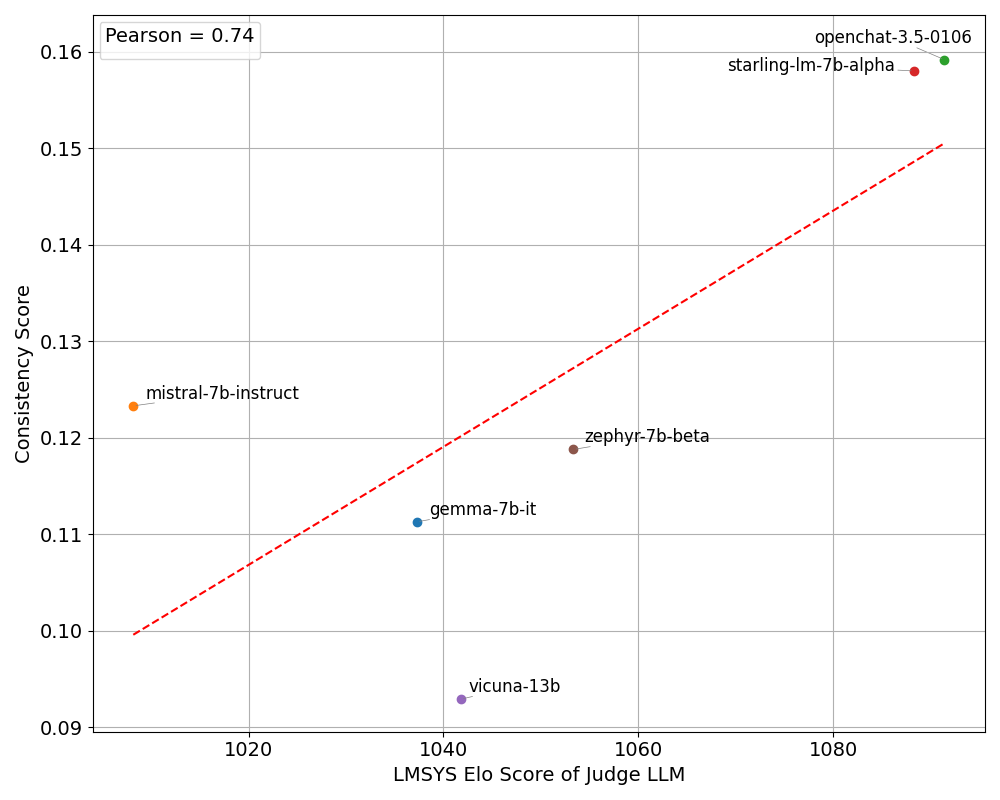}
        \label{fig:correlation_lowcluster}
    }
    \caption{Correlation plots between Consistency and Elo scores for each judge LLM, grouped into clusters of low-, middle-, and high-performing models.}
    \label{fig:correlation_grid}
\end{figure}

\begin{figure}[H]
    \centering
    \includegraphics[width=\linewidth]{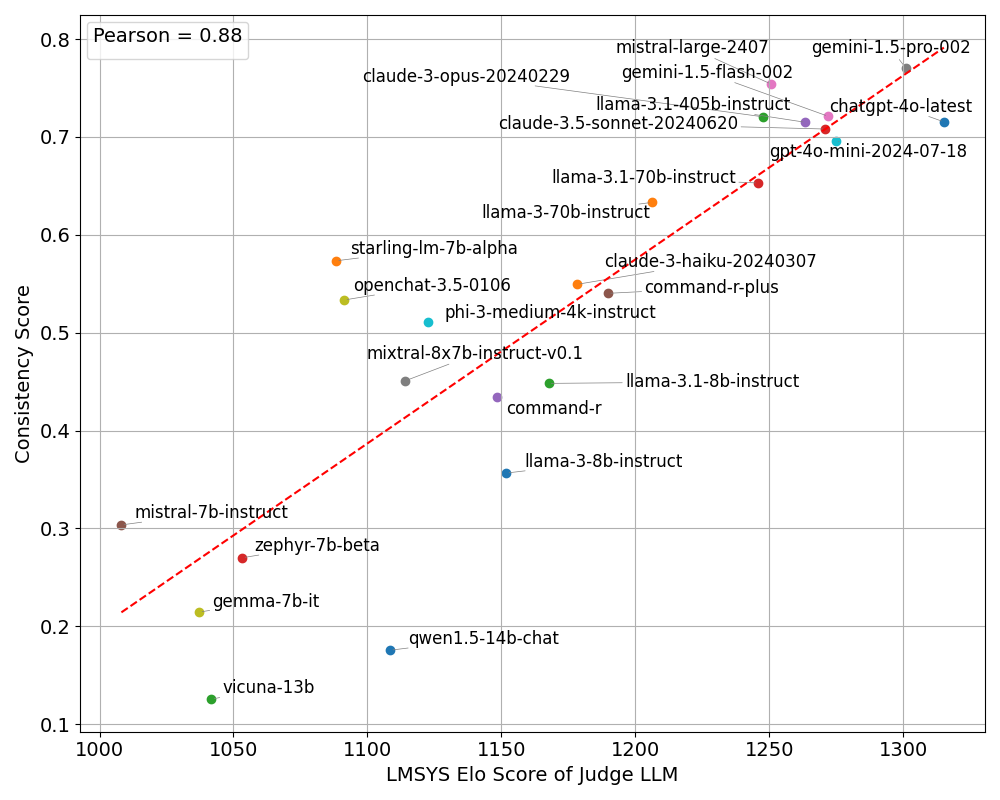}
    \caption{Correlation plot between the Elo and Consistency scores of each judge, computing the Consistency scores using \emph{only} the 10 matchups with the highest Elo differences.}
    \label{fig:correlation_top10matchups}
\end{figure}

\end{document}